\documentclass[journal]{IEEEtran}

\usepackage{booktabs}
\usepackage{multirow}
\usepackage{threeparttable}
\usepackage{makecell}
\usepackage{array}
\usepackage[font=small,labelfont=bf]{caption}
\usepackage{amsmath}
\usepackage{amssymb}
\usepackage{pifont}
\usepackage{colortbl}
\usepackage{graphicx}
\usepackage{siunitx}
\usepackage{url}
\usepackage[table]{xcolor}   

\newcommand{\method}[3]{%
  #1\textsubscript{\textcolor{gray!70}{\scriptsize\itshape #2'#3}}%
}

\newcommand{\best}[1]{\textcolor{red}{\textbf{#1}}}

\newcommand{\second}[1]{\textcolor{blue}{\textbf{#1}}}
\IEEEoverridecommandlockouts
\IEEEaftertitletext{\vspace{-2.5\baselineskip}}

\ifCLASSINFOpdf
\else
\fi
\IEEEpubid{This work has been submitted to the IEEE for possible publication. Copyright may be transferred without notice, after which this version may no longer be accessible.}

\begin{document}
%
\title{Prompt-Calibrated SAM 3 for Open-Vocabulary Remote Sensing Semantic Segmentation}
%
%
%

\author{Yanghui Song, Nanqing Liu, Haonan Yin, Yingjie Gao, Chengfu Yang, and Qi Ming%
\thanks{This work was supported in part by the Basic Research Project of Natural Science of Yunnan Province under Grants 202301AT070065 and 202601AT070017.}
\thanks{Corresponding author: Chengfu Yang (yangchengfu@ynnu.edu.cn).}
\thanks{Yanghui Song (yanghuisong55@gmail.com), Nanqing Liu (lansing163@163.com), Haonan Yin (haonanyin26@gmail.com), and Chengfu Yang (yangchengfu@ynnu.edu.cn) are with the School of Information Science and Technology, Yunnan Normal University, Kunming, China.}%
\thanks{Yingjie Gao (gaoyingjie@buaa.edu.cn) is with the School of Computer Science and Engineering, Beihang University, Beijing, China.}%
\thanks{Q. Ming (chaser.ming@gmail.com) is with the College of Computer Science, Beijing University of Technology, Beijing, China.}}

\maketitle
\IEEEpubidadjcol
\begin{abstract}
Open-vocabulary semantic segmentation (OVSS) in remote sensing images aims to segment categories beyond a fixed label space. Recent SAM 3-based methods provide a promising training-free foundation, yet three key issues remain: (1) a single class-name prompt lacks sufficient semantic coverage for complex remote sensing categories; (2) expanding each category into multiple prompts introduces redundant online text encoding; and (3) directly aggregating multiple prompt responses propagates noisy activations into the final prediction. To address these issues, we propose ProC-SAM3, which calibrates SAM 3's prompt interface for remote sensing OVSS from three complementary aspects. First, we construct an offline prompt pool where a Category Matcher groups MLLM-generated candidates into per-category sets, and Expansion Constraints further refine each set using category-specific prior knowledge. Second, the resulting text embeddings are cached and reused across all test images, eliminating repeated text encoding. Third, we introduce Presence-Guided Residual Fusion to gate unreliable decoder outputs by prompt presence and confidence, followed by peak-preserving class aggregation that retains fine-grained activations for small and sparse objects. Experiments on eight benchmarks show that ProC-SAM3 achieves an average mIoU of 56.1\%, outperforming the previous best training-free method by 3.9 percentage points. Code will be available at \url{https://github.com/YanghuiSong/ProC-SAM3}.
\end{abstract}


%
\section{Introduction}
\label{sec:intro}

Open-vocabulary semantic segmentation (OVSS) in remote sensing images aims to interpret complex geospatial scenes beyond a fixed set of predefined categories. Existing vision-language segmentation methods~\cite{lan2024clearclip,zhang2025corrclip,cao2025open,hu2026sota} leverage CLIP-style patch-level features to transfer category semantics, but their lack of instance-aware segmentation capability limits performance on remote sensing images~\cite{li2025segearth,li2025annotation,li2025segearthov3}. The recent SAM 3, a ViT-based foundation segmentation model, with its built-in promptable mask decoder, offers a fundamentally stronger segmentation backbone and has quickly surpassed CLIP-based methods on remote sensing OVSS benchmarks~\cite{carion2025sam,liu2025pointsam,pei2026taming}.

\begin{figure}[t]
    \centering
    \includegraphics[width=\columnwidth]{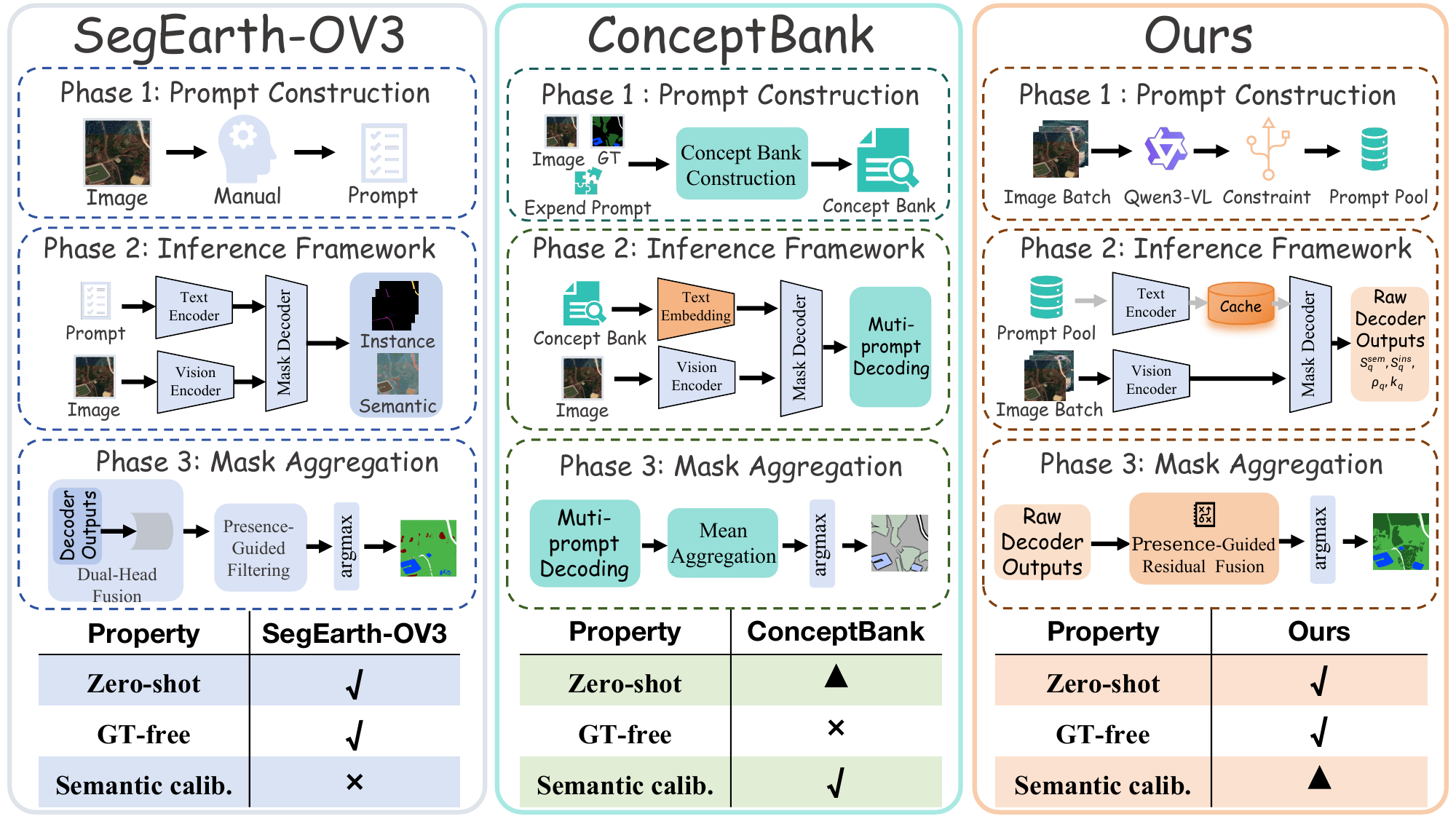}
    \caption{Comparison of SAM 3 adaptation paradigms for remote sensing OVSS across three phases: prompt construction, inference, and mask aggregation. SegEarth-OV3 lacks semantic calibration and encodes prompts repeatedly online; ConceptBank achieves calibration but relies on ground-truth annotations; our ProC-SAM3 provides GT-free calibration with cached embeddings and presence-guided aggregation.}\vspace{-1.0\baselineskip}
    \label{fig:paradigm_comparison}
\end{figure}

However, the key challenge has shifted from segmentation capability to prompt calibration: how should SAM 3's text-prompt interface be adapted to remote sensing categories? Three issues arise in practice: (1) a single class-name prompt provides insufficient semantic coverage for visually diverse remote sensing categories; (2) naively expanding each category into multiple prompts multiplies online text encoding cost; and (3) directly aggregating multiple prompt responses propagates noisy activations into the final prediction. As illustrated in Fig.~\ref{fig:paradigm_comparison}, existing adaptation strategies address these issues to different extents across three phases: prompt construction, inference, and mask aggregation. SegEarth-OV3~\cite{li2025segearthov3} uses manually designed prompts without semantic calibration, encodes all prompts online for every image, and fuses decoder outputs via dual-head fusion with presence-guided filtering. While fully training-free and GT-free, it lacks domain-aware prompt calibration and incurs redundant text encoding. ConceptBank~\cite{pei2026taming} constructs a semantically calibrated concept bank and caches text embeddings to avoid repeated encoding, but its concept bank is built from ground-truth masks, limiting applicability to unannotated scenarios. Moreover, its mean aggregation over multiple prompt responses tends to suppress activations of small objects.

\begin{figure*}[t]
    \centering
    \includegraphics[width=1\linewidth]{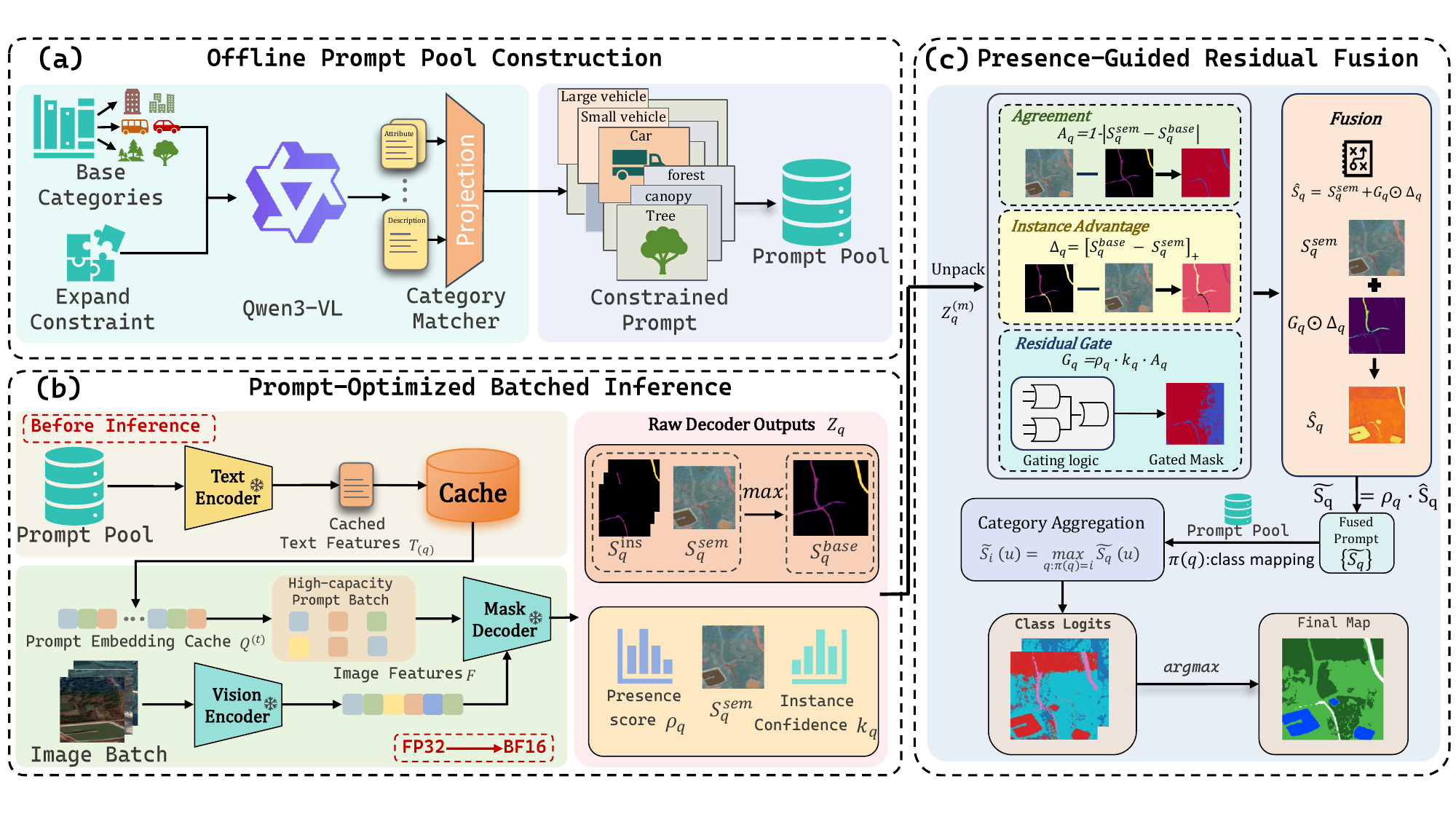}
    \caption{Overview of the proposed ProC-SAM3, which operates in a fully training-free manner through three stages: (a) it first constructs a semantically enriched prompt pool offline via MLLM candidate generation, category matching, and expansion constraint refinement; (b) it then caches the prompt embeddings and performs efficient batched decoding; and (c) it finally fuses per-prompt responses into per-class predictions through presence-guided residual gating and pixel-wise max-pooling aggregation.}\vspace{-1.0\baselineskip}
    \label{fig:mainFig}
\end{figure*}
To this end, we propose ProC-SAM3, a training-free and GT-free framework that improves SAM 3 adaptation for remote sensing OVSS across prompt construction, inference, and mask aggregation. Our contributions are as follows:
\newpage
\begin{itemize}
    \item We design Offline Prompt-pool Construction (OPC) to calibrate category prompts without using ground-truth masks. OPC uses an MLLM to generate candidate descriptions and refines them through category matching and constraint-based projection, producing a compact prompt pool with clearer semantic boundaries.
    \item We introduce Prompt-Optimized Batched Inference (POBI) to reduce the cost introduced by prompt expansion. POBI encodes the offline prompt pool once, caches the text embeddings, and reuses them during batched decoding, avoiding redundant online text encoding.
    \item We propose Presence-Guided Residual Fusion (PGRF) for multi-prompt mask aggregation. PGRF gates per-prompt decoder outputs by presence score and instance confidence, and then applies pixel-wise max-pooling to preserve fine-grained activations for small and sparse objects.
    \item Experiments on eight remote sensing OVSS benchmarks show that ProC-SAM3 achieves an average mIoU of 56.1\%, outperforming the previous best training-free method by 3.9 percentage points.
\end{itemize}

\section{Methodology}

Given a remote sensing image $x \in \mathbb{R}^{H \times W \times 3}$ and a category set $\mathcal{C}=\{c_1,\dots,c_N\}$, our goal is to predict pixel-wise semantic labels $\hat{Y} \in \{1,\dots,N\}^{H \times W}$ without any task-specific training. As illustrated in Fig.~\ref{fig:mainFig}, ProC-SAM3 improves SAM 3's prompt interface from three complementary perspectives: \emph{what to prompt}, where Offline Prompt-pool Construction (OPC) enriches each category into a semantically diverse prompt set (Sec.~\ref{sec:pool}); \emph{how to encode}, where Prompt-Optimized Batched Inference (POBI) caches the prompt embeddings once and reuses them across all test images (Sec.~\ref{sec:infer}); and \emph{how to aggregate}, where Presence-Guided Residual Fusion (PGRF) gates unreliable prompt responses and retains fine-grained activations through pixel-wise max-pooling (Sec.~\ref{sec:pgrf}).
\vspace{-0.5cm}
\subsection{Offline Prompt-pool Construction}
\label{sec:pool}

A single class-name prompt often provides insufficient semantic coverage for complex remote sensing categories, because the same category may appear in different shapes, scales, and scene contexts. To address this, we build an offline prompt pool through a coarse-to-fine semantic refinement process, as illustrated in Fig.~\ref{fig:mainFig}(a). In this process, each base category acts as a semantic anchor, and the final goal is to expand it into a compact set of descriptive prompts that are diverse enough to capture appearance variation but still stable enough to be reused at test time.

We first utilize Qwen3-VL to generate a diverse candidate set by jointly processing each base category $c_i$ and its expansion constraints $EC_i$. The raw outputs are intentionally broad, containing both informative and potentially ambiguous candidates. To resolve ambiguity, we apply the Category Matcher (CM) function, which maps each candidate to its optimal base class. For example, "large vehicle," "small vehicle," and "car" are assigned to the vehicle class, while "forest," "canopy," and "tree" are grouped under trees. This step yields the refined category-specific candidate set $\widetilde{\mathcal{P}}_i$ for each $c_i$, formalized as:
\begin{equation}
\widetilde{\mathcal{P}}_i = \text{CM}(\text{Qwen3}(c_i, EC_i)) .
\end{equation}
The matched candidates are then refined by category-specific expansion constraints, which encode prior knowledge about expected attributes and descriptions. In the figure, these constraints are summarized as Attribute and Description cues. In practice, they can cover scale and material cues, object form, visual appearance, and scene environment, such as ``large vehicle,'' ``forest,'' ``asphalt road,'' or ``paved road.'' Candidates that violate these constraints are discarded, and only descriptions that remain consistent with the intended category semantics are retained. This refinement step plays the role of a lightweight projection: it keeps the useful diversity produced by the MLLM while removing prompts that are semantically loose or visually implausible for the remote sensing domain.
\begin{equation}
    \mathcal{P}_i = \operatorname{Projection}\bigl(\widetilde{\mathcal{P}}_i,\; \text{Attribute}_i,\; \text{Description}_i\bigr) .
\end{equation}

All per-category sets are merged into a global prompt pool $\mathcal{P}^* = \bigcup_{i=1}^{N}\mathcal{P}_i$, where each prompt $q$ retains its category label $\pi(q)$ (with $\pi(q) \in \{1,\dots,N\}$ mapping the prompt to its associated category index). The resulting pool is therefore not just a list of category names, but a curated set of executable text prompts with clearer semantic boundaries. The entire construction is performed offline and adds no cost at test time. Constraint definitions and matching details are provided in our released code\footnote{\url{https://github.com/YanghuiSong/ProC-SAM3}}.
\vspace{-0.5cm}
\subsection{Prompt-Optimized Batched Inference}
\label{sec:infer}

Expanding each category from one prompt to many improves semantic coverage but also multiplies the text encoder calls per image in a conventional online inference pipeline. Since the global prompt pool $\mathcal{P}^*$ is constructed offline and remains fixed for all test images, repeatedly encoding the same textual instructions is unnecessary. We therefore reorganize the inference pipeline into a prompt-optimized batched form (Fig.~\ref{fig:mainFig}(b)). Before online segmentation, the entire prompt pool is mapped into the SAM 3 text feature space once and stored as persistent prompt embeddings:
\begin{equation}
    T_{\{Q\}} = \operatorname{TextEncoder}(\mathcal{P}^*) ,
\end{equation}
where $Q$ denotes the set of prompt indices. We use $q$ to index an individual prompt and $\pi(q)\in\{1,\dots,N\}$ to denote its associated category (consistent with the definition in Sec.~\ref{sec:pool}). This cache is independent of the image content and can be directly reused across datasets or image batches as long as the category set and prompt pool are unchanged.

During online inference, an image batch $\mathcal{I}$ is processed by the vision encoder once to obtain shared visual features $F = \operatorname{VisionEncoder}(\mathcal{I})$. The cached prompt embeddings are split into chunks of size $B$. For the $t$-th chunk, $T_{\mathcal{Q}^{(t)}} \subset T_{\{Q\}}$, the shared visual features and cached prompt features are jointly fed into the mask decoder:
\begin{equation}
    Z_q = \operatorname{MaskDecoder}\bigl(F,\; T_q\bigr), \quad q \in \mathcal{Q}^{(t)} .
\end{equation}
This formulation avoids repeated text encoding and enables batched decoding over multiple prompts and images.

For each prompt $q$, the SAM 3 decoder produces four outputs after Sigmoid activation: semantic mask logits $S_q^{\mathrm{sem}}$, instance mask logits $S_q^{\mathrm{ins}}$, a presence score $\rho_q$ indicating whether the prompted object exists in the image, and an instance confidence $k_q$. Among the instance masks, we select the one with the highest confidence $k_q$ as the representative instance output $S_q^{\mathrm{ins}}$, and define a base response $S_q^{\mathrm{base}}(u) = \max\bigl(S_q^{\mathrm{sem}}(u),\, S_q^{\mathrm{ins}}(u)\bigr)$ that retains the stronger activation at each pixel $u$. These outputs serve as inputs to the subsequent fusion stage.
\vspace{-0.5cm}
\subsection{Presence-Guided Residual Fusion}
\label{sec:pgrf}

After decoding, each category has multiple prompt-level responses that must be combined into a single per-class prediction. Two problems arise in this step: mismatched prompts may produce confident masks on incorrect regions, and simply averaging responses across prompts tends to suppress local activations of small or narrow targets. To address both, we design PGRF as follows. The SAM 3 decoder provides both a semantic branch and an instance branch for each prompt. When a prompt is genuinely present in the image and both branches agree, the instance branch can provide sharper boundary details that complement the semantic response. However, when a prompt is absent or the two branches disagree, the instance response is unreliable and should be suppressed. PGRF implements this logic through three quantities:
\begin{align}
    A_q &= 1 - \left| S_q^{\mathrm{sem}} - S_q^{\mathrm{base}} \right| , \\
    \Delta_q &= \left[ S_q^{\mathrm{base}} - S_q^{\mathrm{sem}} \right]_+ , \\
    G_q &= \rho_q \cdot k_q \cdot A_q .
\end{align}
Here, $A_q$ measures pixel-wise agreement between the semantic and instance branches (values near $1$ indicate high agreement), $\Delta_q$ captures the positive residual where the instance response exceeds the semantic response, and $G_q$ is a gating factor that combines presence $\rho_q$, confidence $k_q$, and agreement $A_q$. The fused prompt-level response is:
\begin{equation}
    \widetilde{S}_q = \rho_q \cdot \left(S_q^{\mathrm{sem}} + G_q \odot \Delta_q \right) .
\end{equation}
When the gate conditions are not met, $G_q$ approaches zero and the output reduces to $\rho_q \cdot S_q^{\mathrm{sem}}$. After fusion, the refined responses are mapped back to their categories via $\pi(q)$ and aggregated by pixel-wise max-pooling:
\begin{equation}
    \widetilde{S}_i(u) = \max_{q:\,\pi(q)=i} \widetilde{S}_q(u) .
\end{equation}
This retains the strongest local activation within each class, avoiding the suppression of small-object responses caused by averaging. Finally, an $\arg\max$ across categories yields the prediction map $\hat{Y}(u) = \arg\max_{i} \widetilde{S}_i(u)$.

\begin{table*}[t]
\centering
\caption{Quantitative comparison on eight remote sensing OVSS benchmarks. OEM denotes OpenEarthMap; $^*$ denotes reproduced results, all of which are evaluated using the \textbf{SegEarth-OV3 class definitions}. All values are mIoU (\%), and Avg. is computed over the reported entries.}
\label{tab:ovss_rs}
\setlength{\tabcolsep}{3.8pt}
\footnotesize

\begin{tabular}{l|c|cccccccc|c}
\toprule
\multirow{2}{*}{\textbf{Method}} & \multirow{2}{*}{\textbf{Backbone}} & \multicolumn{8}{c|}{\textbf{Dataset}} & \multirow{2}{*}{\textbf{Avg.}} \\
\cmidrule(lr){3-10}
 & & \textbf{LoveDA} & \textbf{Potsdam} & \textbf{Vaihingen} & \textbf{iSAID} & \textbf{OEM} & \textbf{UDD5} & \textbf{VDD} & \textbf{UAVid} & \\
\midrule

\method{VIP$^*$}{ICML}{26} \cite{zhu2026vip} & DINOv3 + dino.txt & 31.2 & 23.0 & 30.5 & 23.5 & 20.7 & 35.2 & 52.5 & 19.9 & 29.6 \\

\method{SCLIP}{ECCV}{24} \cite{wang2024sclip}          & CLIP & 30.4 & 39.6 & 35.9 & 16.1 & 29.3 & 38.7 & 37.9 & 31.4 & 32.4 \\

\method{ClearCLIP}{ECCV}{24} \cite{lan2024clearclip}  & CLIP & 32.4 & 42.0 & 36.2 & 18.2 & 31.0 & 41.8 & 39.3 & 36.2 & 34.6 \\

\method{ProxyCLIP}{ECCV}{24} \cite{lan2024proxyclip}  & CLIP+DINOv2 & 34.3 & 49.0 & 47.5 & 21.8 & 38.9 & 40.8 & 47.8 & 35.8 & 39.5 \\

\method{SegEarth-OV}{CVPR}{25} \cite{li2025segearth}  & CLIP & 36.9 & 48.5 & 29.1 & 21.7 & 40.3 & 50.6 & 45.3 & 42.5 & 39.2 \\

\method{RSCLIP$^*$}{J-STARS}{26} \cite{wang2026rsclip} & CLIP & 35.5 & 47.4 & 40.3 & 21.2 & 39.8 & 50.4 & 42.5 & 42.5 & 39.9 \\

\method{CorrCLIP}{ICCV}{25} \cite{zhang2025corrclip}  & CLIP+DINO+SAM2 & 36.9 & 51.9 & 47.0 & 25.5 & 32.9 & 46.1 & 47.3 & 38.3 & 40.7 \\

\method{ConInfer}{CVPR Fingding}{26} \cite{chen2026coninfer}  & CLIP+DINOv3 & 39.3 & 50.0 & 31.4 & 20.1 & \second{42.0} & 46.9 & 50.3 & 46.4 & 40.8 \\

\method{ReAttnCLIP}{CVPR}{26} \cite{niu2026reattnclip}  & CLIP & 37.0 & 48.7 & 29.9 & 23.2 & 41.1 & 53.7 & 49.7 & 44.0 & 40.9 \\

\method{ActiveSAM$^*$}{arXiv}{26} \cite{tien2026activesam} & SAM3 & 41.1 & 40.6 & 45.1 & \second{28.3} & 40.7 & \second{72.1} & 62.2 & 53.3 & 47.9 \\

\method{SegEarth-OV3$^*$}{arXiv}{25} \cite{li2025segearthov3} & SAM 3 & \second{41.6} & \second{57.2} & \second{60.8} & 26.0 & 41.0 & 71.6 & \second{64.4} & \second{54.7} & \second{52.2} \\

\textbf{Ours} & SAM 3 & \best{46.1} & \best{58.8} & \best{64.5} & \best{29.5} & \best{48.9} & \best{73.4} & \best{67.5} & \best{60.0} & \best{56.1} \\

\bottomrule
\end{tabular}
\end{table*}

\section{Experiments}
\subsection{Implementation Details}
We evaluate on eight remote sensing datasets: Vaihingen, Potsdam, LoveDA, OpenEarthMap, iSAID, UAVid, UDD5, and VDD. Input images are resized to $1008 \times 1008$ by the SAM3 processor. The offline prompt pool is constructed once per dataset using Qwen3-VL-8B under constraint files, and text embeddings are pre-computed and cached by the prompt bank. During inference, prompts are processed in batches with a size of $1$ for UDD5, VDD, and UAVid, and $4$ for other datasets. The image encoder operates in half precision while the decoder remains in full precision. All experiments are conducted without training on a single RTX 4090 GPU (24GB), with Qwen3 unloaded from GPU memory before SAM3 inference to avoid memory contention. Unless otherwise specified, quantitative results are reported as mIoU (\%).

\begin{figure}[t]
    \centering
    \includegraphics[width=\columnwidth]{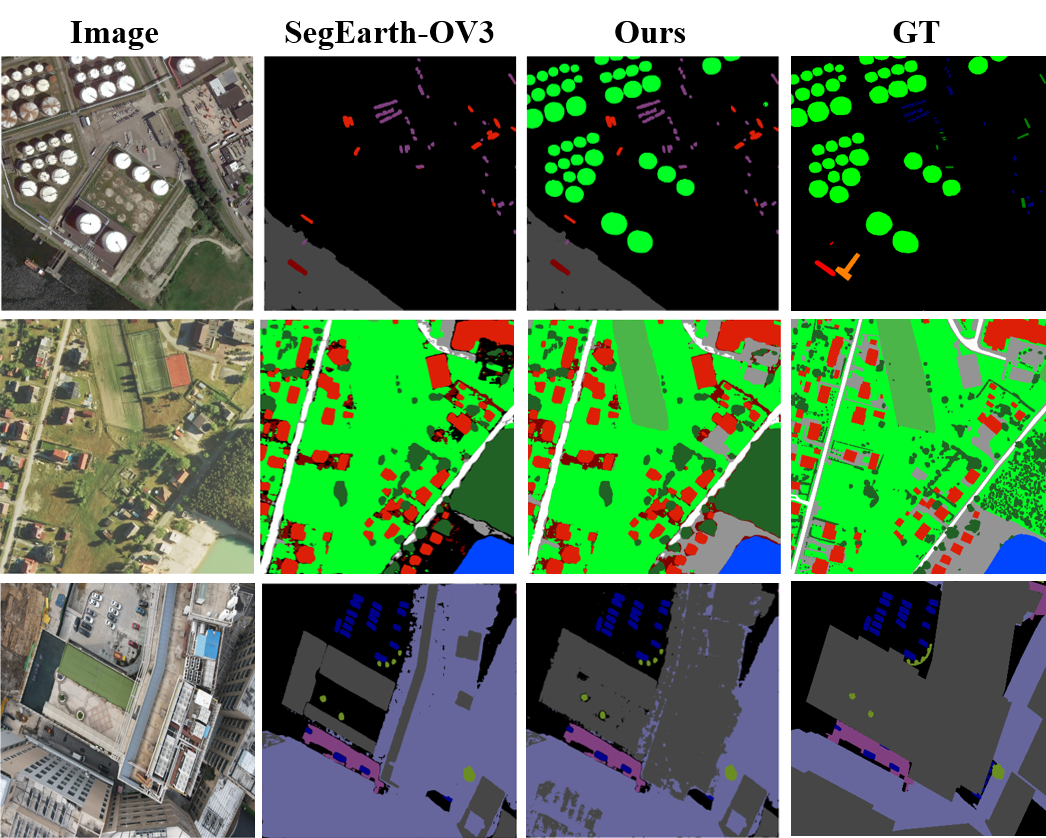}
    \caption{Qualitative comparison on representative remote sensing scenes.}
    \label{fig:Segmentation}
    \vspace{-0.5cm}
\end{figure}
\vspace{-0.5cm}
\subsection{Comparison with State-of-the-Art Methods}
Table~\ref{tab:ovss_rs} compares representative open-vocabulary segmentation methods on eight remote sensing benchmarks. Since several prior methods do not report results on all datasets and the average is computed over available entries, the fairest direct comparison is with training-free methods that provide full coverage, especially SegEarth-OV3. Under this setting, our method achieves the highest mIoU on all eight benchmarks and improves the average mIoU from 52.2 to 56.1 (+3.9). The gains are consistent across datasets and are most notable on OpenEarthMap (+7.9\%), UAVid (+5.3\%), LoveDA (+4.5\%), and Vaihingen (+3.7\%). These datasets contain diverse land-cover patterns, dense small objects, and large appearance variations, suggesting that calibrated prompts are more effective than directly using category names or unconstrained prompt expansion.

The qualitative results in Fig.~\ref{fig:Segmentation} show a similar trend. Compared with SegEarth-OV3, our method recovers more complete structures, generates cleaner land-cover regions, and retains stronger small-object responses. These improvements indicate that offline image-conditioned prompt calibration and presence-guided mask aggregation reduce semantic ambiguity while preserving fine-grained activations in complex remote sensing scenes.
\vspace{-0.5cm}
\subsection{Ablation Study}

\subsubsection{Component Ablation}

\begin{table}[t]
\centering
\caption{Ablation study of framework components on four remote sensing datasets. PC denotes Prompt Calibration and PGRF denotes Prompt-Guided Residual Fusion.}
\label{tab:ablation}
\setlength{\tabcolsep}{3.2pt}
\footnotesize
\begin{tabular}{cc|ccccc}
\toprule
\textbf{OPC} & \textbf{PGRF} & \textbf{UDD5}  & \textbf{OEM} & \textbf{LoveDA} & \textbf{Avg.} \\
\midrule
 &  & 71.6 & 41.0  & 41.6 & 51.4 \\
\ding{51} &  & 72.8  & 48.3 & 44.3 & 55.1 \\
\ding{51} & \ding{51} & \textbf{73.4} & \textbf{48.9} & \textbf{46.1} & \textbf{56.1} \\
\bottomrule
\end{tabular}
\end{table}

\begin{figure}[t]
    \centering
    \begin{minipage}{0.95\linewidth}  
        \begin{minipage}[t]{0.48\columnwidth}
            \centering
            \includegraphics[width=\linewidth]{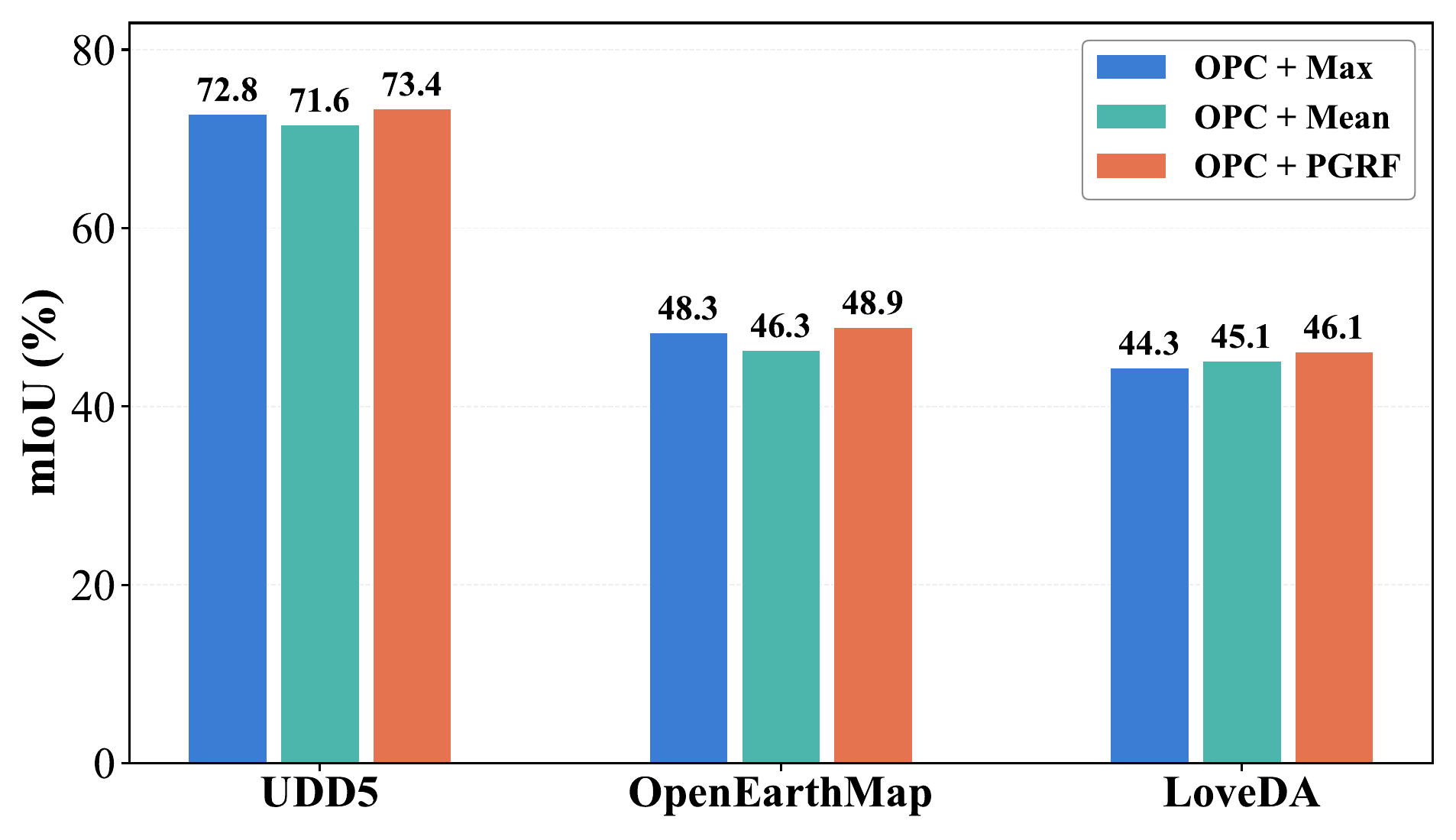}
            \caption{Ablation on different fusion strategies.}\vspace{-1.0\baselineskip}
            \label{fig:Abl}
        \end{minipage}
        \hfill
        \begin{minipage}[t]{0.48\columnwidth}
            \centering
            \includegraphics[width=\linewidth]{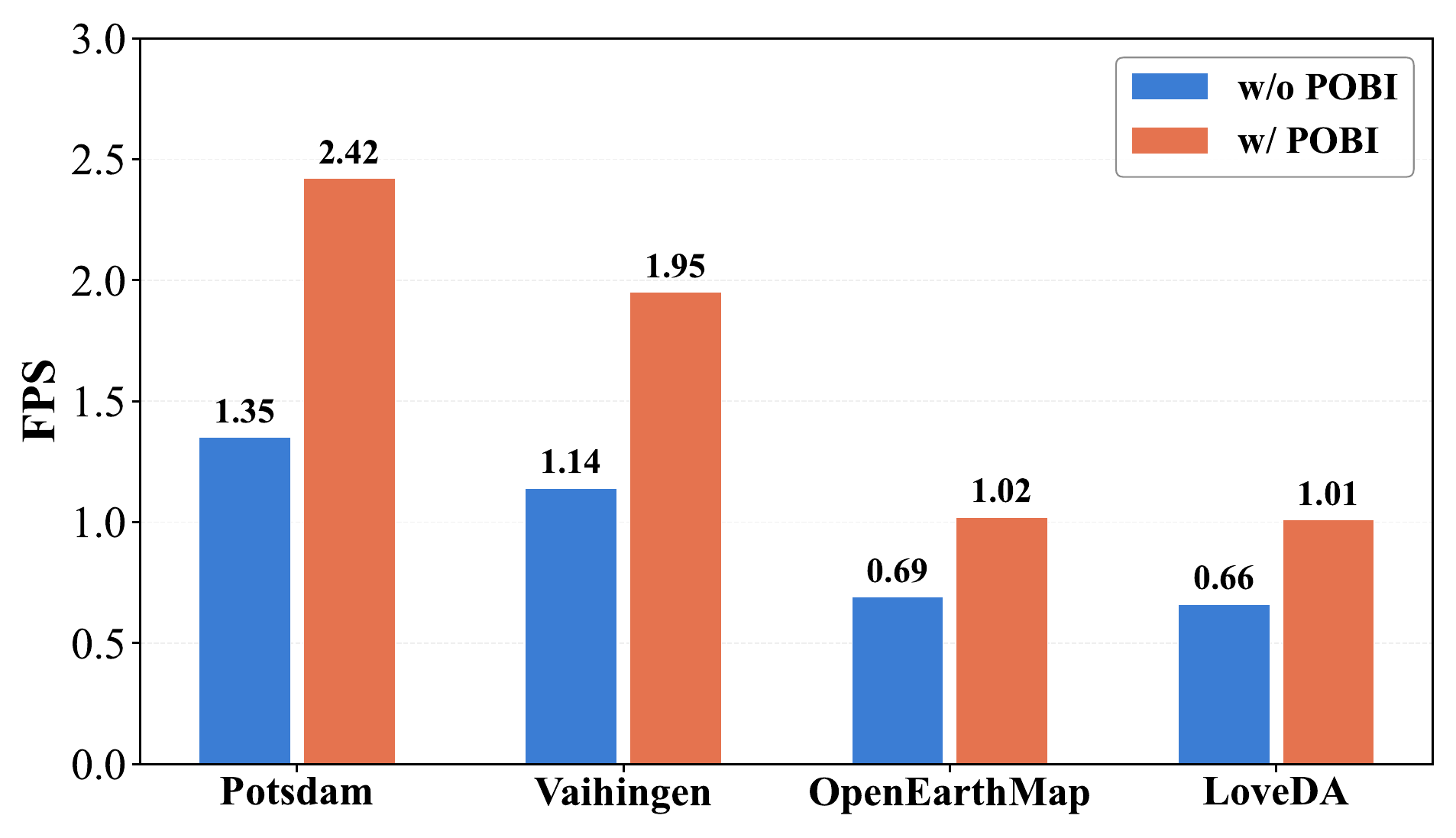}
            \caption{Inference speed comparison with and without POBI.}\vspace{-1.0\baselineskip}
            \label{fig:FPS}
        \end{minipage}
    \end{minipage}%
\end{figure}

As shown in Table~\ref{tab:ablation}, OPC provides the dominant gain (+3.7\% avg. mIoU), indicating the baseline bottleneck is semantic alignment rather than decoder capacity. PGRF further improves the average to 56.1 with consistent gains across all datasets. Fig.~\ref{fig:Abl} extends this analysis by comparing PGRF against two alternative fusion strategies (max and mean) across three datasets. Max and mean achieve comparable performance (55.1 vs.~54.4), while PGRF outperforms both, confirming its effectiveness in suppressing unreliable instance activations. Fig.~\ref{fig:FPS} shows that POBI consistently accelerates inference across all datasets, with the largest gain observed on Potsdam (1.35$\rightarrow$2.42 FPS).

\begin{table}[htbp]
\centering
\setlength{\tabcolsep}{4pt}
\caption{Ablation on prompt construction strategies across three datasets. All variants use the same inference pipeline.}
\label{tab:prompt_expansion_compact}
\begin{tabular}{lccccc}
\toprule
\textbf{Prompt Strategy} & \textbf{UDD5} & \textbf{OEM} & \textbf{LoveDA} & \textbf{Avg.} \\
\midrule
Class-name prompt & 71.6 & 41.0 & 41.6 & 51.4 \\
Open-ended MLLM prompt & 66.4 & 31.3 & 34.6 & 44.1 \\
Text-only expansion & 72.3 & 40.6 & 42.8 & 51.9 \\
Image-conditioned expansion & 72.7 & 45.1 & 44.3 & 54.0 \\
Constraint-guided prompt pool & \textbf{73.4} & \textbf{48.9} & \textbf{46.1} & \textbf{56.1} \\
\bottomrule
\end{tabular}
\vspace{2pt}
\end{table}

\subsubsection{Prompt Construction Strategies}
We further evaluate how different prompt construction strategies affect semantic alignment. Table~\ref{tab:prompt_expansion_compact} compares five strategies under the same inference pipeline: class-name prompts, open-ended MLLM prompts, text-only expansion, image-conditioned expansion, and the proposed constraint-guided prompt pool. To make the difference more concrete, we take the \emph{road} class in UDD5 as an example. The class-name baseline uses only ``\emph{road}.'' The open-ended MLLM setting often produces broad variants such as ``\emph{road}, \emph{street}, \emph{path}, \emph{highway}, \emph{avenue},'' while text-only expansion narrows them to words such as ``\emph{road}, \emph{path}, \emph{paved}.'' After introducing image context, the image-conditioned expansion yields more scene-aware phrases such as ``\emph{asphalt road}, \emph{paved road}, \emph{school road}.'' Our constraint-guided prompt pool further filters these candidates and retains a stable subset, e.g., ``\emph{road}, \emph{asphalt road}, \emph{paved road},'' thereby preserving prompt diversity without causing semantic drift.

The quantitative results in Table~\ref{tab:prompt_expansion_compact} are consistent with the above examples. Open-ended MLLM prompts degrade performance on all three datasets, dropping the average mIoU from 51.4 to 44.1, which indicates that unconstrained expansion introduces substantial semantic drift. Text-only expansion is more stable and slightly improves the average to 51.9, but its gains remain limited because lexical variation alone cannot reliably capture remote sensing scene context. After introducing image context, the average mIoU further rises to 54.0, with a particularly clear gain on OEM (40.6 $\rightarrow$ 45.1), showing that scene-aware prompt generation is already beneficial. The proposed constraint-guided prompt pool achieves the best results on all three datasets, reaching 73.4 on UDD5, 48.9 on OEM, and 46.1 on LoveDA. Compared with the class-name baseline, this corresponds to gains of 1.8, 7.9, and 4.5 points, respectively, indicating that the combination of image conditioning and category-specific constraints yields the most reliable prompt set.

\subsubsection{Efficiency of Prompt-Optimized Batched Inference}

Fig.~\ref{fig:FPS} compares the inference throughput without and with POBI. The figure shows consistent gains on all four datasets, from 1.14 to 1.95 on Vaihingen, 0.69 to 1.02 on OpenEarthMap, 0.66 to 1.01 on LoveDA, and 1.35 to 2.42 on Potsdam. This gain mainly comes from avoiding repeated text encoding and replacing fp32 inference with bf16, which reduces both computation and memory overhead during decoding.

\section{Conclusion}
We presented ProC-SAM3, a training-free framework that improves SAM 3's prompt interface for remote sensing OVSS from three aspects: prompt semantics (OPC), encoding efficiency (POBI), and mask aggregation (PGRF). Experiments on eight benchmarks show consistent improvements over existing methods, with an average mIoU of 56.1\%. Ablation results indicate that OPC contributes the largest accuracy gain, while PGRF and POBI provide complementary improvements in prediction quality and inference efficiency, respectively.



\bibliographystyle{IEEEtran}
\bibliography{references}

%




\end{document}